\documentclass[10pt,twocolumn]{article}

\usepackage[T1]{fontenc}
\usepackage[utf8]{inputenc}
\usepackage[USenglish]{babel}
\usepackage[a4paper,margin=1in]{geometry}
\usepackage{times}
\usepackage{hyperref}
\usepackage{graphicx}
\usepackage{booktabs}
\usepackage{authblk}
\usepackage[
  backend=biber,
  style=numeric,
  sorting=none,
  doi=false,
  url=false,
  isbn=false,
  eprint=false,
  giveninits=true,
  maxnames=3,
  minnames=1
]{biblatex}

\AtEveryBibitem{
  \clearfield{month}
  \clearfield{day}
  \clearfield{note}
  \clearfield{issn}
}

\title{Data Synthesis Improves 3D Myotube Instance Segmentation}
\author{David Exler$^{1}$, Nils Friederich$^{1,2}$, Martin Kr\"uger$^{1}$, John Jbeily$^{3}$, Mario Vitacolonna$^{3}$, R\"udiger Rudolf$^{3}$, Ralf Mikut$^{1}$, Markus Reischl$^{1}$}

\date{}

\begin{document}

\maketitle
\begin{center}
\small
Submitted to BMT 2026 (VDE). This is a preprint and has not yet been peer-reviewed.
\end{center}
\begingroup
\renewcommand{\thefootnote}{\arabic{footnote}}
\footnotetext[1]{Institute for Automation and Applied Informatics, Karlsruhe Institute of Technology, Hermann-von-Helmholtz-Platz 1, 76344 Eggenstein-Leopoldshafen, Germany.}
\footnotetext[2]{Institute of Biological and Chemical Systems, Karlsruhe Institute of Technology, Hermann-von-Helmholtz-Platz 1, 76344 Eggenstein-Leopoldshafen, Germany.}
\footnotetext[3]{CeMOS Research and Transfer Center, Technische Hochschule Mannheim, Paul-Wittsack-Straße 10, 68163 Mannheim, Germany.}
\footnotetext[0]{Corresponding author: David Exler, david.exler@kit.edu}
\endgroup

\begin{abstract}
    Myotubes are multinucleated muscle fibers serving as key model systems for studying muscle physiology, disease mechanisms, and drug responses. Mechanistic studies and drug screening thereby rely on quantitative morphological readouts such as diameter, length, and branching degree, which in turn require precise three-dimensional instance segmentation. Yet established pretrained biomedical segmentation models fail to generalize to this domain due to the absence of large annotated myotube datasets. We introduce a geometry-driven synthesis pipeline that models individual myotubes via polynomial centerlines, locally varying radii, branching structures, and ellipsoidal end caps derived from real microscopy observations. Synthetic volumes are rendered with realistic noise, optical artifacts, and CycleGAN-based Domain Adaptation (DA). A compact 3D U-Net with self-supervised encoder pretraining, trained exclusively on synthetic data, achieves a mean IPQ of 0.22 on real data, significantly outperforming three established zero-shot segmentation models, demonstrating that biophysics-driven synthesis enables effective instance segmentation in annotation-scarce biomedical domains.
\end{abstract}


\section{Introduction}
In this paper, we show that existing state-of-the-art biomedical segmentation models, such as CellposeSAM~\cite{pachitariu_cellpose-sam_2025}, PlantSeg~\cite{wolny_accurate_2020}, and StarDist~\cite{weigert2020}, fail to generalize to 3D myotube instance segmentation. 
These models are predominantly trained on compact, roughly spherical cell morphologies and lack exposure to the highly elongated, branching, and mutually overlapping geometry of myotubes. 
Manual annotation of 3D myotube data is particularly challenging: individual instances can span hundreds of voxels in length, frequently touch or cross neighboring fibers, and exhibit locally varying thickness and ellipsoidal nuclear bulges, making unambiguous delineation extremely laborious even for domain experts. 
Recent research aims to alleviate annotation efforts by introducing data synthesis pipelines \cite{bohland2023unsupervised, bruch2025improving, bruch_synthesis_2023}. 
As no large annotated 3D myotube datasets, nor a myotube simulation pipeline exist, we tackle this challenge by presenting the first data synthesis pipeline for 3D myotube data and training exclusively on synthetic data. 
We validate our approach by benchmarking a compact U-Net~\cite{ronneberger_u-net_2015} with a self-supervised learning (SSL) pretrained encoder against the three established models, contributing a foundation for scalable analysis of myotubes as key model systems in muscle development, disease modeling, and drug screening~\cite{Gilbert2014, lewis1917behavior, Park2024Dexamethasone, Weisrock2024MyoFInDer}.
\section{Methods}
\subsection{Synthesis}
\begin{figure}
	\includegraphics[width=\columnwidth]{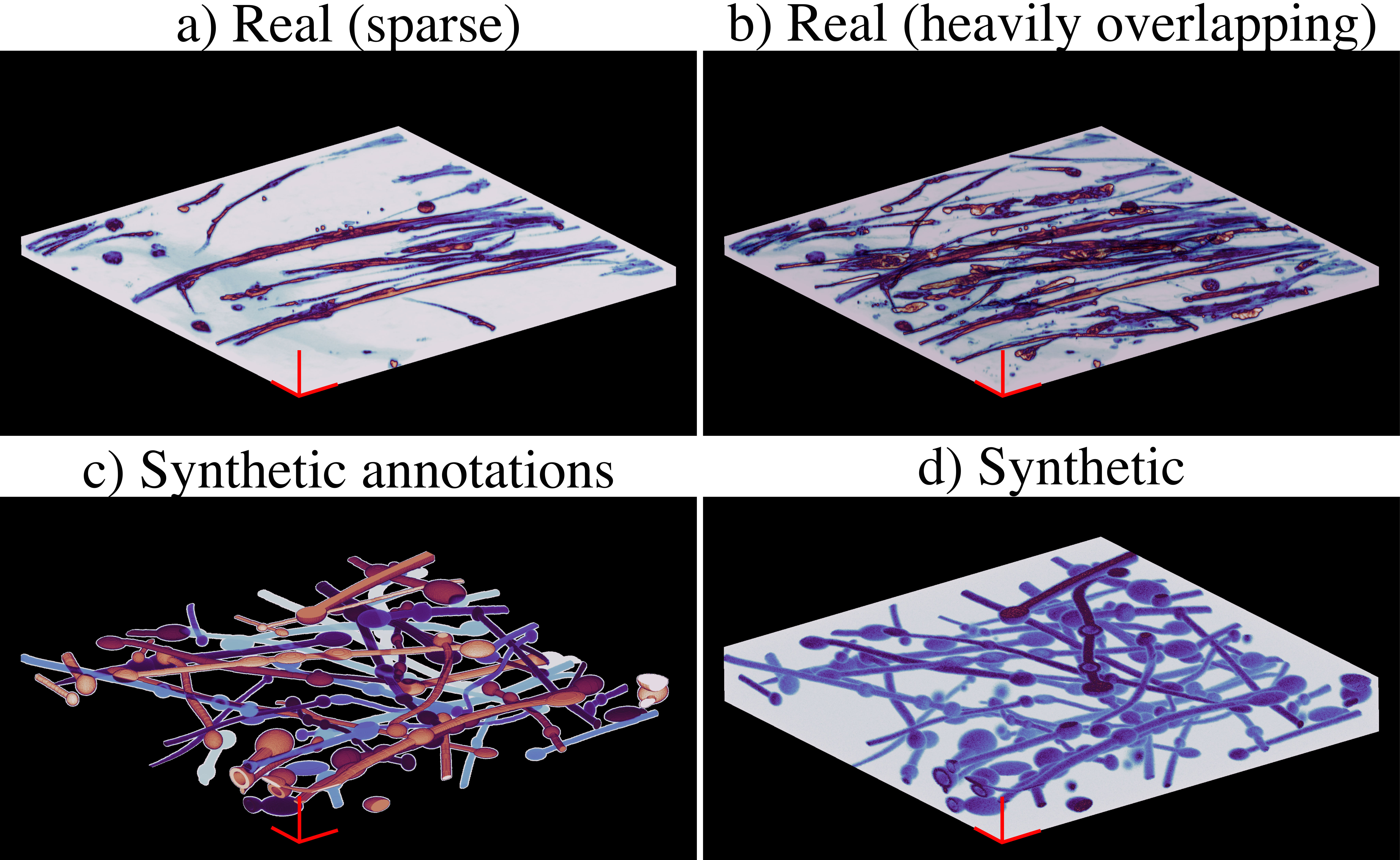}
	\caption{3D renderings illustrating the synthesis pipeline. 
Real myotube data ranges from sparsely distributed (a) to 
heavily overlapping configurations (b). The pipeline generates 
instance-labeled annotation volumes (c) and corresponding 
synthetic fluorescence image stacks (d) that closely 
approximate real imaging conditions.}
	\label{img:Figure1}
\end{figure}

We derive geometrical characteristics from real myotube data (see Figs.~\ref{img:Figure1}a, b) 
and generate synthetic annotation volumes in three stages. First, polynomial centerlines 
are sampled in XY and Z using damped Chebyshev basis functions $T_k$ of degree $K$ with 
coefficients $\tilde{c}_k \sim \mathcal{U}(-1,1)$ scaled by $k^{-\alpha}$ to suppress 
boundary oscillations. Local thickness is modulated by a smooth polynomial and a sinusoidal 
term controlled by $\gamma$ and $\delta$. Second, straight branching segments are inserted 
while preserving slope continuity. Third, ellipsoidal structures are placed along the 
centerline, with hollow ellipsoids along the shaft and solid caps at endpoints. Realistic 
fluorescence appearance is simulated by additive Poisson and Gaussian noise, salt-and-pepper 
textures, debris artifacts, anisotropic halos, and Gaussian PSF blurring 
(Figs.~\ref{img:Figure1}c, d). 
A CycleGAN is trained on unpaired real and synthetic images to further generate domain-adapted images~\cite{bruch_synthesis_2023}.

\subsection{Segmentation Models}
We train two compact residual 3D U-Nets with four encoder stages, each comprising two convolutional layers, a residual connection, max pooling, and normalization. 
The decoder progressively upsamples via transposed convolutions with skip connections and residual refinement blocks.\\
Because domain-specific SSL pretraining can greatly benefit segmentation models, the encoder of one model is pretrained on real myotube data using Fully Convolutional Masked Autoencoding (FCMAE)~\cite{woo2023convnext}: 3D input patches are randomly masked, and the network learns to reconstruct the missing regions. 
This SSL variant is extended with additional encoder layers beyond the four shared stages, which remain trainable during downstream optimization. 
The SSL pretrained encoder layers are frozen, while the additional layers and the entire second model are trained from scratch. 
We train the segmentation decoder twice for each model, once on unadapted synthetic data and once on the data we domain-adapt using CycleGAN.
This results in four model variants: (1) UNet, (2) UNet + DA, (3) SSL-pretrained UNet, and (4) SSL-pretrained UNet + DA.
Two output channels are learned to predict the foreground and centerline probabilities.
Thresholding the foreground probability predicts a binary mask and a seeded Watershed algorithm with the thresholded centerline probability separates the instances.
We benchmark our model against three established zero-shot biomedical segmentation models 
(see Table \ref{tab:segm_models}).
\begin{table}[h!]
    \caption{Segmentation models evaluated in this study. 
Three established zero-shot biomedical segmentation models 
are compared against our models trained 
exclusively on synthetic data.}
    \begin{tabular}{llr}
        \textbf{Model} & \textbf{Abbrev.} & \textbf{Param.} \\
        \midrule
        CellposeSAM~\cite{pachitariu_cellpose-sam_2025} & CSAM & 304.6M \\
        PlantSeg~\cite{wolny_accurate_2020}             & psG  & 4M     \\
        StarDist~\cite{weigert2020}                     & SD   & 1.4M   \\
        U-Net (ours)                                    & UNet & 5M    \\
        U-Net + DA (ours)                   & UNet (DA) & 5M    \\
        SSL U-Net (ours)                                & SSL  & 23M    \\
        SSL U-Net + DA (ours)               & SSL (DA) & 23M    \\
    \end{tabular}
    \label{tab:segm_models}
\end{table}

\subsection{Dataset}
We generate 200 individual synthetic instances across 30 images of 128 x 1024 x 1024 pixels for our training dataset. 
To validate our synthesis pipeline on real data, we 
manually annotate a subset of a 3D myotube fluorescence microscopy 
dataset.
This dataset contains 17 images of shape (z, 1024, 1024), with z ranging from 24 to 128~\cite{Couturier2024}. 
Due to the challenges described above, we perform sparse annotation, labeling $n=40$ individual myotube instances across 17 volumetric stacks. 
\\\\Instances are selected to cover a representative range of morphological 
configurations, including straight, curved, and 
branching myotubes, as well as varying degrees of 
neighbor overlap. These annotations serve exclusively 
as an independent test set; no annotated data is 
used during training. 

\section{Results \& Discussion}

\subsection{Visual Inspection}

\begin{figure*}[]
    \centering
	\includegraphics[width=0.9\linewidth]{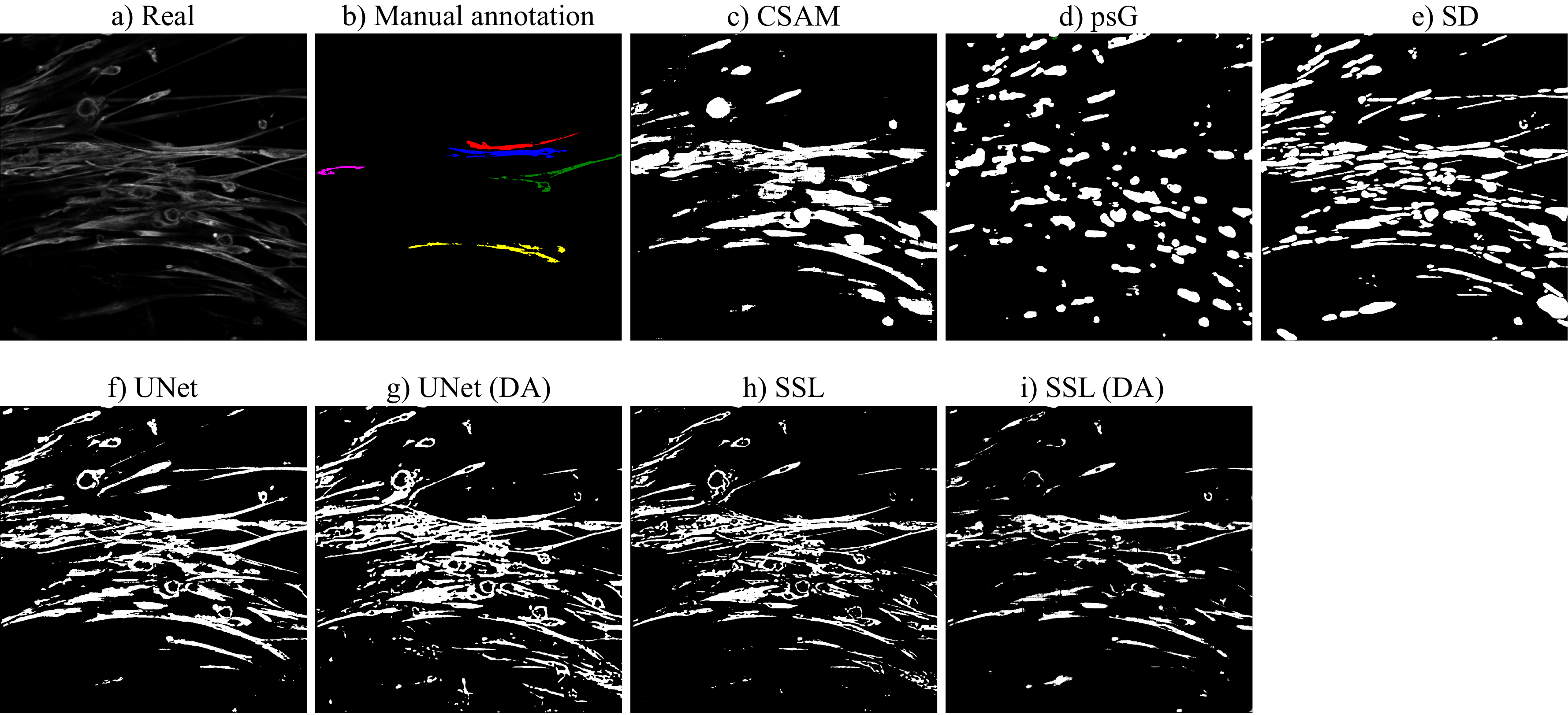}
	\caption{Examples of 2D slices. a): Real, b): Corresponding sparse manual annotation, c)-e): Segmentation masks predicted by the pretrained segmentation models, and f)-i): Segmentation masks by our models.}
	\label{img:Figure2}
\end{figure*}

\begin{figure}[h]
    \centering
	\includegraphics[width=0.9\linewidth]{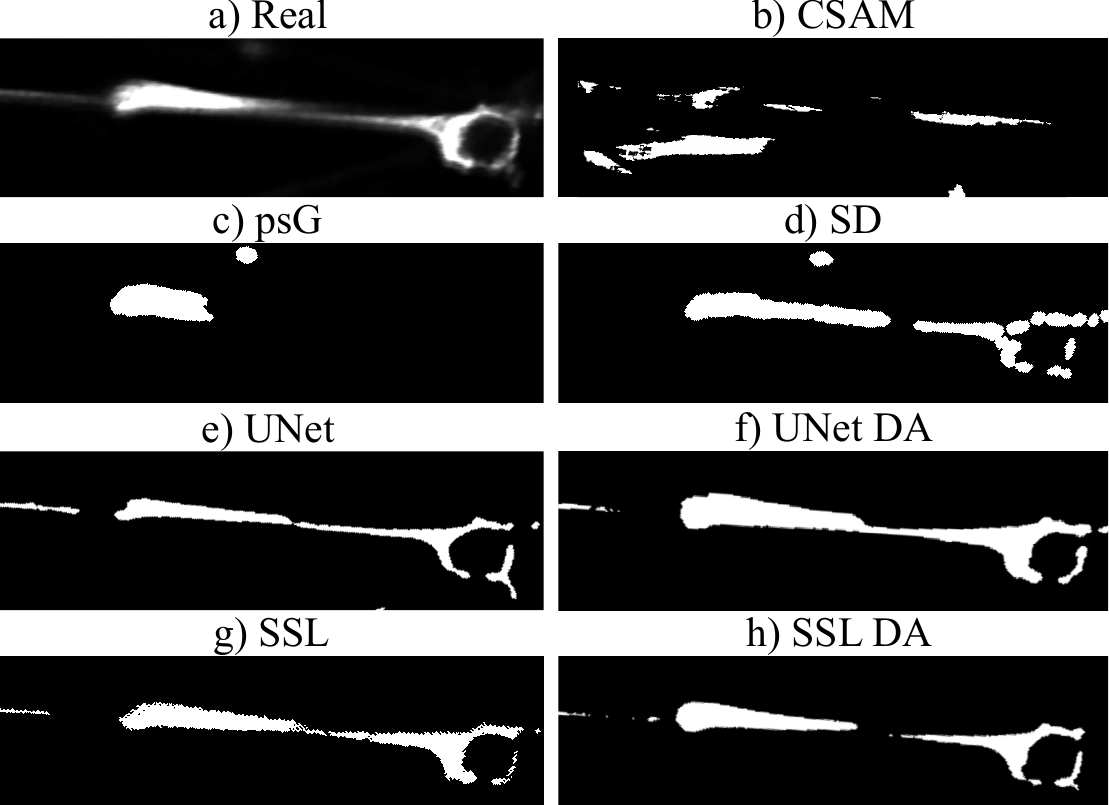}
	\caption{Examples of zoomed-in 2D slices. a): Real, b)-h): Segmentation masks predicted for this instance by the segmentation models included in this study.}
	\label{img:Figure3}
\end{figure}

Fig.~\ref{img:Figure2} shows a representative 2D slice of 
the real dataset (a) alongside the sparse manual annotations (b) of this slice and the predictions from all models. 
CSAM (c) merges multiple instances into large irregular blobs and responds strongly to background halo artifacts, failing to resolve individual myotubes. 
psG (d) segments numerous small round regions, reflecting its training bias 
toward compact cell morphologies. 
SD (e) similarly produces fragmented oval pieces without capturing the elongated myotube geometry. 
In contrast, our models (f-i) successfully detect the elongated structures, though it partially predicts shadow artifacts as foreground. 
The domain-adapted UNet (g) further reduces these false positives, producing cleaner foreground redictions that match the manual annotations (b) more closely.
Fig.~\ref{img:Figure3} provides a closer inspection of a single myotube instance featuring a thin elongated body, a pronounced nuclear bulge, and a tapering distal end.CSAM (b) produces disconnected fragments at incorrect locations. 
psG (c) detects only the bright nuclear bulge while entirely missing the elongated body. 
SD (d) splits the instance into three separate fragments, recovering parts of the geometry but lacking continuity along the myotube axis. 
Both our UNet models (e-f) correctly capture both the elongated shaft and the nuclear bulge, with only the thin distal tip partially missed. 
\\The SSL prediction (g) also captures these relevant features, but is noisy on the edges.
The domain-adapted SSL (h) produces the closest match to the real instance, preserving the overall shape, including the nuclear bulge with clean boundaries. 
\subsection{Quantitative Comparison}
We evaluate all models using the Injective Panoptic Quality (IPQ)~\cite{exler2026}, which jointly measures segmentation accuracy and instance detection while penalizing incorrect splits, with values ranging from 0 to 1.
Since only a subset of instances is annotated, the recall component is disabled. 
Standard metrics such as Dice or PQ assume complete annotations and would unfairly penalize models for detecting unannotated instances, making IPQ the appropriate choice here. 
Fig.~\ref{img:Figure4} reports the mean IPQ with standard error over $n=40$ annotated instances, with significance assessed via paired t-tests across instances.
\begin{figure}[h]
	\includegraphics[width=\linewidth]{IPQ.pdf}
	\caption{Mean IPQ scores with standard error of the mean over $n=40$ annotated instances for all evaluated models. 
    Significance of pairwise differences between our best model (SSL (DA)) and each baseline was assessed via paired t-tests (*** $p < 0.001$). 
    Our models (marked with *) consistently outperform all three zero-shot baselines despite having far fewer parameters. 
    DA denotes training on domain-adapted synthetic data.}
	\label{img:Figure4}
\end{figure}
\\All of our models score significantly higher than all three zero-shot baselines ($p < 0.001$), despite having far fewer parameters than CSAM.
Our domain-adapted SSL model achieves a mean IPQ of 0.22, demonstrating the substantial impact of our data synthesis pipeline.
The remaining gap to a perfect score reflects the domain shift between synthetic training data and real imaging conditions, which we identify as the primary direction for future improvement.

\subsection{Ablation Study}
Because the results of Fig.~\ref{img:Figure4} show large performance gaps between the SSL (DA) model (0.22) and our other models (0.034 - 0.039), we disentangle the contributions of DA and SSL pretraining, and we compare all four model variants (Table \ref{tab:segm_models}). 
SSL pretraining without using domain-adapted data to learn the downstream task yields a marginal decline (0.039 to 0.038), likely because the pretrained features are aligned with the real data domain, while the downstream training on unadapted synthetic data introduces a domain mismatch that limits effective transfer.
In contrast, combining SSL with DA leads to a substantial performance gain (0.22), outperforming all other variants by a large margin.
Without a strong feature initialization, this additional appearance complexity can hinder learning of the underlying geometry and increase the risk of overfitting to GAN-specific artifacts, explaining the marginal performance drop in UNet (DA) relative to UNet. 
However, when combined with SSL pretraining, the encoder already provides robust real-domain feature representations, allowing the model to effectively exploit the reduced domain gap introduced by DA. This interaction explains the large performance gain observed exclusively in the SSL (DA) combination.

\section{Conclusion}
We present the first geometry-driven synthesis pipeline for 3D myotube data, enabling instance segmentation in a domain where no annotated training data exists. 
A compact SSL-pretrained U-Net trained exclusively on synthetic data significantly outperforms three established zero-shot biomedical segmentation models, demonstrating that controlled geometry-driven synthesis can substitute for costly manual annotations. 
The approach generalizes to similar annotation-scarce domains such as 3D nerve fiber or mycelium networks.
Future work will focus on closing the remaining domain gap, incorporating disease-specific morphological variations, and scaling toward a large, well-performing instance segmentation model for myotube data. The synthesis pipeline is publicly available at \href{https://github.com/DavidExler/syn_myo}{github.com/DavidExler/syn\_myo}.\\

\textsf{\textbf{Author Statement}}\\
Research funding: The Helmholtz Association funds this project under the research school "Helmholtz Information and Data Science School for Health (HIDSS4Health)", and the program "Natural, Artificial and Cognitive Information Processing (NACIP)". RR was funded by the Baden-Württemberg-Stiftung, grant LSE-012, and the DFG, grant INST874/9-1. Conflict of interest: Authors state no conflict of interest. Informed consent: Informed consent has been obtained from all individuals included in this study. Ethical approval: Does not apply.

\printbibliography[heading=bibintoc,title={References}]

\end{document}